\documentclass[11pt]{article}

\usepackage[margin=1in]{geometry}
\usepackage{booktabs}
\usepackage{float}
\usepackage{graphicx}
\usepackage{hyperref}
\usepackage{microtype}
\usepackage[numbers]{natbib}

\newcommand{\samethanks}[1][\value{footnote}]{\footnotemark[#1]}

\title{A Negative Result on Cross-Model Activation Transfer in a Pythia Multi-Hop Setting}
\author{Peiyan Zhang\thanks{These authors contributed equally to this work (Peiyan Zhang, Jason Xin).} \\
Independent Researcher \\
\texttt{zhangpeiyan2004@gmail.com}
\and
Jason Xin\samethanks \\
Independent Researcher \\
\texttt{xinjingshun@foxmail.com}}
\date{}

\begin{document}
\maketitle

\begin{abstract}
Recent work shows that language models can transmit behavioural traits through
hidden signals in generated data during training. We ask whether a different
activation-mediated channel is viable: can one language model communicate a
useful intermediate reasoning state to another at inference time through a
post-hoc linear activation bridge, rather than through a textual or
structured-token relay? We test this question in a controlled Pythia-160M to
Pythia-410M multi-hop reasoning setting. A linear translation
layer learns a strong normalized-space map between sender and receiver hidden
states, with normalized cosine similarity near 0.97 across seeds. However,
when the translated activations are injected into the receiver at inference
time, they do not improve downstream answering. Low-strength additive
injection remains near the no-injection baseline, with confidence intervals
that cross zero. Replacement-style injection is consistently destructive, and
rescaling translated vectors to the receiver hidden-state norm does not rescue
performance. The result is therefore a scoped negative result: in this setting,
offline representational alignment is not sufficient for useful causal
communication inside the receiver.
\end{abstract}

\section{Introduction}

How does hidden information move from one language model to another, and which
channels actually carry it? Recent work demonstrates one surprising channel: a
model can transmit behavioural traits---such as preferences or
misalignment---to another model through hidden signals in generated training
data, even when explicit references to the trait are filtered out
\citep{cloud2026subliminal}. That transfer is data-mediated, occurs at training
time through fine-tuning or distillation, operates at the level of a model-wide
trait, and is observed primarily when the sender and receiver share the same or
a behaviourally matched base model. This report tests a different
activation-mediated channel within the same broad question. Instead of routing
a signal through generated data and a training step, we translate a sender's
hidden activations and inject them into a receiver at inference time, asking
whether a single instance-level reasoning state can transfer through a
post-hoc linear bridge without any adaptation of the receiver.

Activation-level communication is attractive because it promises to bypass a
serialization bottleneck. In many current multi-agent large-language-model
systems, information passed between models is mediated by text, structured
messages, tool outputs, memory objects, or other token-level interfaces, which
still require a sender to externalize internal state and a receiver to
reconstruct useful internal representations from that relay. If a sender model
has built a useful internal representation, a receiver might in principle
benefit from consuming that representation directly instead of reading a
textual or structured-token relay. This activation-transfer study tests this
idea in a deliberately narrow setting: Pythia-160M sends hidden states to
Pythia-410M through a learned linear translation layer, and the receiver uses
the translated states during multi-hop question answering.

The result is negative. The translation layer is not trivially bad; it fits the
normalized receiver activation space well. The failure appears when the
translated states are used causally inside the receiver. This distinction is
the core contribution of the report. A vector can be aligned under an offline
objective while still being unusable as a replacement for the receiver's own
hidden state trajectory.

This report makes three claims. First, the tested activation-transfer
mechanism does not outperform either no-injection or natural-language relay,
even though the translation layer reaches normalized $R^2 \approx 0.88$ and
normalized cosine similarity $\approx 0.97$ across seeds; the contrast between
high offline alignment and zero downstream improvement is a clean separation
of representational alignment from receiver-side causal usability. Second,
replacement-style transfer is actively harmful, and the failure decomposes
into two quantifiable factors: a roughly two-orders-of-magnitude mismatch in
L2 norm between the translated vector and the receiver's native hidden state,
and a residual direction or distribution error in the translated state that
remains destructive even after the L2-norm gap is closed by scale correction.
Third, the experiment narrows the design problem for future work: useful
model-to-model activation communication likely requires a training objective
that optimizes receiver-side causal use, not only sender-to-receiver
representational fit.

We use \emph{activation transfer} to mean inference-time insertion of hidden
states derived from one model into another. The \emph{sender} is Pythia-160M,
the \emph{receiver} is Pythia-410M, and the \emph{translation layer} is a
learned linear map between their hidden dimensions. The
\emph{natural-language relay} baseline lets the sender generate text and gives
that text to the receiver. \emph{Replacement} injection substitutes the
receiver hidden state at the injection layer with translated sender
activations, while \emph{additive} injection adds a small translated vector to
the receiver hidden state. The claim is intentionally scoped to one model
family, one sender-receiver pair, one task family, one fixed injection depth,
and one linear translation mechanism; it should not be read as a general
impossibility claim about activation communication, activation steering,
model stitching, or representation engineering.

\section{Related Work}

The experiment sits at the intersection of representation alignment,
cross-model transfer, and causal intervention.

Model stitching connects part of one network to part of another through a
learned mapping and asks whether the two halves compose into a working system
\citep{lenc2015understanding,bansal2021revisiting,hernandez2023model}. Most
prior stitching work targets vision models. Recent work has begun to extend
the framework to language models: \citet{chen2025transferring} use affine
maps to transfer sparse-autoencoder features across language models of
different scales and report that small and large models learn substantially
compatible feature spaces. That line of work is still organized around
offline feature transferability. The question this report asks is strictly
stronger: whether a translated activation can serve as a usable intermediate
computation inside the receiver at inference time.

Representation similarity work, including CKA \citep{kornblith2019similarity},
has established that independently trained networks often learn statistically
compatible internal spaces. These methods are diagnostic. They certify
structural compatibility but do not test whether one model's hidden state can
serve as a causally usable intermediate computation for another. The negative
result here is consistent with high CKA-style similarity yet still
incompatible with causal substitution.

Activation steering and representation engineering show that directions in
activation space can causally affect generation
\citep{turner2023steering,zou2023representation}. These interventions operate
within a single model: the steering vector is drawn from the model's own
activation distribution and the perturbation is typically a small additive
nudge. Cross-model insertion through a translation layer is strictly harder,
both because the source distribution differs and because replacement injection
is a much stronger intervention than additive steering.

Causal-intervention work inside a single model establishes that mid-layer
activations carry a strong causal role in transformer computation:
activation patching localizes circuits and information flow
\citep{wang2022ioi,zhang2023activationpatching}, induction-head analysis
traces in-context learning to specific attention components
\citep{olsson2022incontext}, and direct knowledge editing modifies factual
associations through targeted parameter updates
\citep{meng2022rome,meng2023memit}. Subsequent editing work tightens the
precision required for non-disruptive intervention:
\citet{fang2025alphaedit} project parameter updates onto a null space of
preserved knowledge to bound interference with unrelated capabilities, and
\citet{tian2024knowundo,zhang2024oneedit} extend editing to selective
unlearning and to neural-symbolic settings. The recurring lesson across this
body of work is that even a same-model substitute activation must remain
close to the state it replaces; the translated cross-model state produced
here is, by construction, much further away.

Lens and embedding-space analyses
\citep{belrose2023tunedlens,dar2023embedding} support the related observation
that internal activations are decodable into interpretable output-space
information. Decodability, however, is weaker than causal substitutability,
which is the property tested here.

A separate line of work in multi-agent reinforcement learning has explored
differentiable latent communication channels, in which agents are jointly
trained so that one produces signals the other consumes. Such methods rely
on shared optimization and co-evolving protocols. Many current multi-agent
large-language-model systems, by contrast, route model-to-model information
through serialized text, structured messages, tool results, or memory objects
rather than shared latent states. The present setup differs from both regimes:
two independently trained language models, no joint training, no receiver
adaptation, and a single post-hoc linear bridge trained offline.

A complementary line of work shows that hidden information can move between
models without any explicit channel at all. \citet{cloud2026subliminal} report
subliminal learning: a teacher model with some trait generates data---number
sequences, code, or chain-of-thought traces---on prompts unrelated to that
trait, and a student fine-tuned on the filtered data acquires the trait, even
after references to it are rigorously removed. This establishes that
cross-model trait transfer is possible, but through a channel that differs from
the one tested here along three axes. Their channel is data-mediated, whereas
ours is activation-mediated; it operates at training time through fine-tuning,
whereas ours operates at inference time with no receiver training; and it
transfers a model-wide behavioural trait, whereas we attempt to transfer a
single instance-level reasoning state. Their effect also appears primarily when
teacher and student share the same or a behaviourally matched base model, which
suggests that successful hidden transfer depends on receiver compatibility and
adaptation, not only on the existence of a hidden signal.

In summary, prior work shows either that internal representations of
different models are structurally similar, or that activations inside a single
model can be causally manipulated. The conjunction---whether high cross-model
offline representational similarity transfers into receiver-side causal
usability under inference-time injection---is what this report tests
directly. Set against the data-mediated, training-time channel of subliminal
learning, the present negative result is best read as a counterexample along
the channel dimension rather than as evidence against hidden model-to-model
transfer in general. The result is compatible with the lines of work above but
imposes a stricter standard. It is not enough that a representation is decodable or
predictable. For model-to-model activation communication, the receiver must
be able to use the inserted state as part of its own computation.

\section{Methods}
\label{sec:methods}

\begin{figure}[H]
\centering
\includegraphics[width=\linewidth]{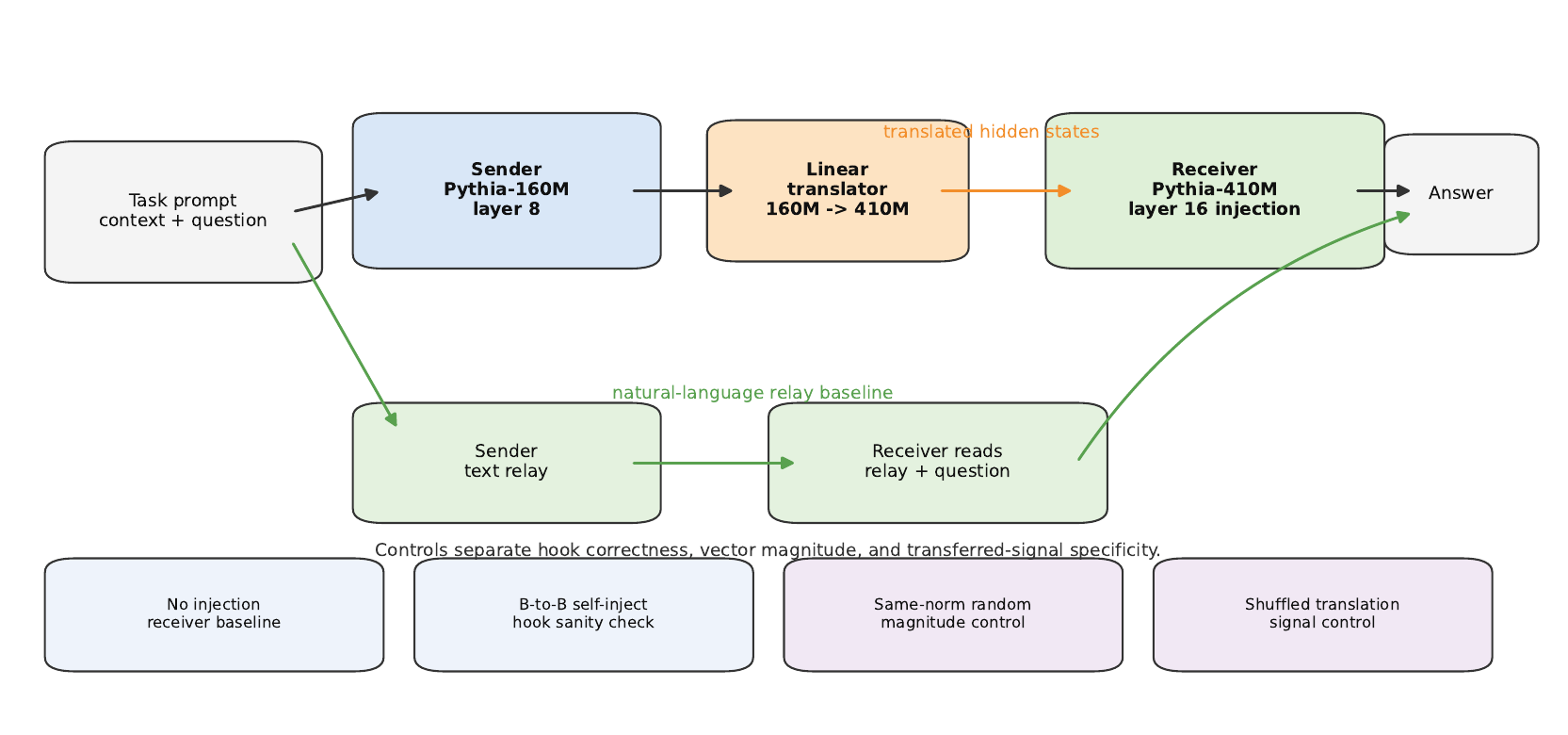}
\caption{Activation-transfer method overview. The primary activation-transfer
path maps sender hidden states through a linear translator and injects them
at the receiver's fixed layer. The relay baseline and controls use the same
clean evaluation set but test different communication or diagnostic channels.}
\label{fig:method}
\end{figure}

Figure~\ref{fig:method} summarizes the experimental path. The sender and
receiver both process the task prompt when activations are extracted and
injected. A separate natural-language relay baseline lets the sender transmit
text instead of hidden states. Injection conditions vary from no-injection
through low-strength additive perturbation to full replacement, so that scale
mismatch and direction or distribution error can be separated in the downstream analysis.

\subsection{Task and Models}

The primary direction is Pythia-160M to Pythia-410M
\citep{biderman2023pythia}. Both models use the Pythia-family GPT-NeoX
tokenizer. The injection depth is fixed at relative depth 0.67, corresponding
to sender layer \texttt{gpt\_neox.layers.8} and receiver layer
\texttt{gpt\_neox.layers.16}. The layer is fixed before the final clean rerun,
so the reported result is not a post-hoc best-layer search.

Pythia is chosen to eliminate confounders that are unrelated to the
activation-transfer question. Sender and receiver share the GPT-NeoX
architecture and the same tokenizer, so identical inputs produce identical
token sequences, and sequence-level injection is well defined; sender and
receiver hidden dimensions differ, so a non-trivial cross-model translation
problem remains.

The task is multi-hop reasoning. Each example contains a context, a question,
and a short gold answer. The final clean evaluation set has 396 rows. Its
frozen SHA-256 hash is:

\begin{quote}
\texttt{504e077cf17433e22967c86e98d32153}\\
\texttt{2d4e803dbe24d96af14c7e8ecdd0dcbb}
\end{quote}

The hash is recorded in every final result JSON and checked by the validation
script.

\subsection{Activation Translation}

The translation layer is a linear map from sender hidden dimension to receiver
hidden dimension. It is trained against L2-normalized receiver activations. In
the primary forward direction, the corrected M6 translation runs show strong
normalized-space alignment, with mean normalized $R^2=0.882773$ and mean
normalized cosine similarity $0.973526$ across seeds.

\subsection{Injection Protocol}

For activation-transfer and no-injection conditions, the receiver uses the full
task prompt containing the source context and question. The sender receives the
same prompt when hidden states are extracted. This makes sequence alignment a
hard protocol requirement: replacement-style injection is only valid when
sender and receiver prompt token sequences have matching length and token IDs.
The runner records both checks for every sample.

For natural-language relay, the sender first generates relay text from the
source context. The receiver then answers from the relay text plus the
question. This baseline is not identical to the activation-transfer input
format, but it is the relevant ordinary communication baseline: one model can
communicate information to another through text.

Receiver decoding is greedy. It uses 64 new tokens, repetition penalty 1.3,
and a no-repeat 3-gram constraint. The natural-language sender relay uses 128
new tokens. These values are serialized into result files and checked during
validation.

The final clean rerun includes no-injection, natural-language relay, additive
injection, uncorrected replacement, scale-corrected replacement, a historical
best-alpha variant with $\alpha=0.30$, B-to-B self-injection, same-norm random
control, zero replacement, full-n shuffled translation, and strict matched
shuffled translation.
The strict matched shuffled control evaluates 383 samples because it drops
singleton prompt-length buckets to avoid self-fallback. The distinction
between the two shuffled controls matters for interpretation: the full-n
shuffled condition preserves the 396-row evaluation set but has 13 fallback
samples per seed because some prompt-length buckets contain only one item,
while the strict matched shuffled condition drops those singleton buckets and
is the paper-primary shuffled diagnostic.

\subsection{Evaluation Metrics}

Exact match is reported but is not the main interpretive metric because the raw
models rarely emit only the short gold answer. The primary lenient metric is
word-boundary answer containment. Historical substring contains-match is
reported for continuity but is treated as too permissive for the main claim.
The metric roles are deliberately separated: exact match is strict, word-boundary
containment is the main lenient paper metric, and legacy contains-match is an
audit target rather than primary evidence.

Word-boundary containment normalizes case, whitespace, and common punctuation.
It then checks whether the gold answer appears as a complete word or phrase
inside the prediction. This avoids common substring artifacts, such as counting
a gold answer that appears only as a substring of another word. Legacy contains
uses the historical substring rule and is included for comparison with earlier
project logs, not as the main metric.

Each condition is compared to no-injection and natural-language relay on shared
sample identifiers. Group-level confidence intervals pool paired sample-level
deltas across runs and use a percentile bootstrap interval. For three-seed
full-n conditions, this yields 1188 paired deltas; for the strict matched
shuffled control, it yields 1149 paired deltas. The intervals are not BCa
intervals; they are reported as a transparent nonparametric uncertainty summary
for paired deltas.

\section{Experiments}

We organize the experiments in three layers. Section~\ref{sec:main-results}
establishes that activation transfer fails to produce a stable downstream
gain. Section~\ref{sec:failure} characterizes the failure modes through an
ablation chain that separates L2-norm mismatch from residual direction or distribution error.
Section~\ref{sec:controls} rules out trivial explanations through diagnostic
controls.

\subsection{Main Results}
\label{sec:main-results}

\begin{table}[H]
\centering
\small
\resizebox{\linewidth}{!}{%
\begin{tabular}{lrrrr}
\toprule
Condition & Runs & $n$ & Boundary contains & Legacy contains \\
\midrule
No injection & 1 & 396 & 0.0884 & 0.1035 \\
Natural-language relay & 1 & 396 & 0.0833 & 0.0909 \\
Additive (highest mean) & 3 & 396 & $0.0926 \pm 0.0073$ & $0.1094 \pm 0.0081$ \\
Replace & 3 & 396 & $0.0025 \pm 0.0025$ & $0.0101 \pm 0.0044$ \\
Scale corrected & 3 & 396 & $0.0076 \pm 0.0025$ & $0.0143 \pm 0.0053$ \\
Best alpha & 3 & 396 & $0.0488 \pm 0.0119$ & $0.0715 \pm 0.0127$ \\
B-to-B self-inject & 1 & 396 & 0.0884 & 0.1035 \\
Same-norm random & 3 & 396 & $0.0000 \pm 0.0000$ & $0.0008 \pm 0.0015$ \\
Zero replacement & 1 & 396 & 0.0025 & 0.0025 \\
Shuffled translation & 3 & 396 & $0.0059 \pm 0.0039$ & $0.0093 \pm 0.0053$ \\
Strict matched shuffled & 3 & 383 & $0.0052 \pm 0.0045$ & $0.0087 \pm 0.0060$ \\
\bottomrule
\end{tabular}
}
\caption{Final clean-eval summary. Values are run means with standard
deviations for multi-seed conditions. Normalized exact match is zero for every
condition and is omitted from the compact table. Additive injection has the
highest mean word-boundary containment among activation-transfer conditions
($0.0926$), but its paired confidence interval versus no-injection crosses
zero (Section~\ref{sec:main-results} and Appendix~\ref{app:pairwise}) and
should not be interpreted as a stable improvement.}
\label{tab:main-results}
\end{table}

The two reference conditions are not clearly separated from each other:
no-injection is higher than natural-language relay by $0.0051$ in
word-boundary containment, with paired bootstrap 95\% CI
$[-0.0328,+0.0429]$.

\begin{figure}[H]
\centering
\includegraphics[width=\linewidth]{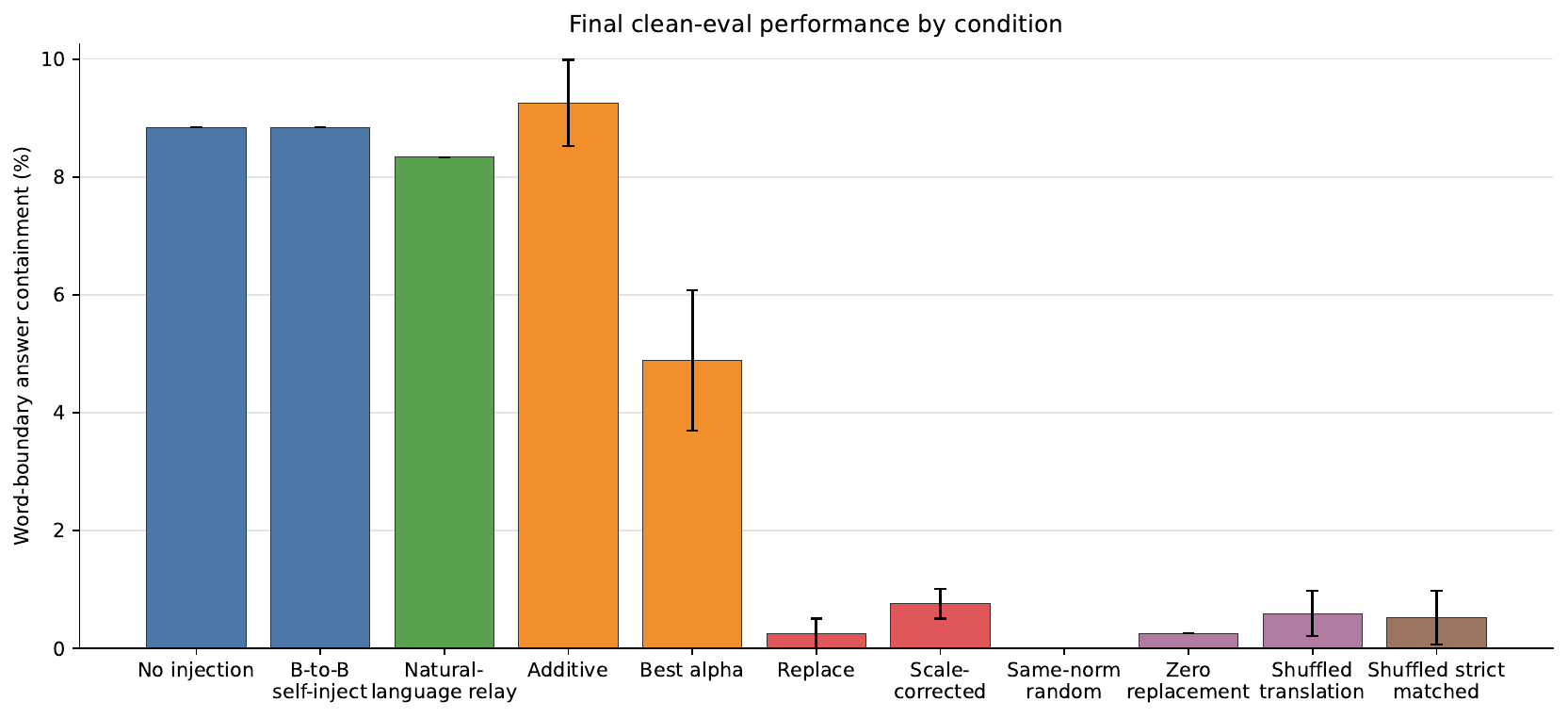}
\caption{Word-boundary answer containment by condition on the locked clean-eval
set. Low-strength additive injection is near the no-injection baseline, while
replacement-style transfer conditions are far below baseline.}
\label{fig:main-boundary}
\end{figure}

The main table and Figure~\ref{fig:main-boundary} show the central pattern.
Additive injection sits near baseline but is not clearly better. Replacement,
scale-corrected replacement, same-norm random injection, and shuffled
translation are all strongly below no-injection and natural-language relay.
Zero replacement is also destructive, giving a lower-bound control for
replacement with no donor direction.
The B-to-B self-injection control exactly matches no-injection, which
indicates that the hook path itself preserves behavior when the inserted
states are native receiver states.

\begin{figure}[H]
\centering
\includegraphics[width=0.88\linewidth]{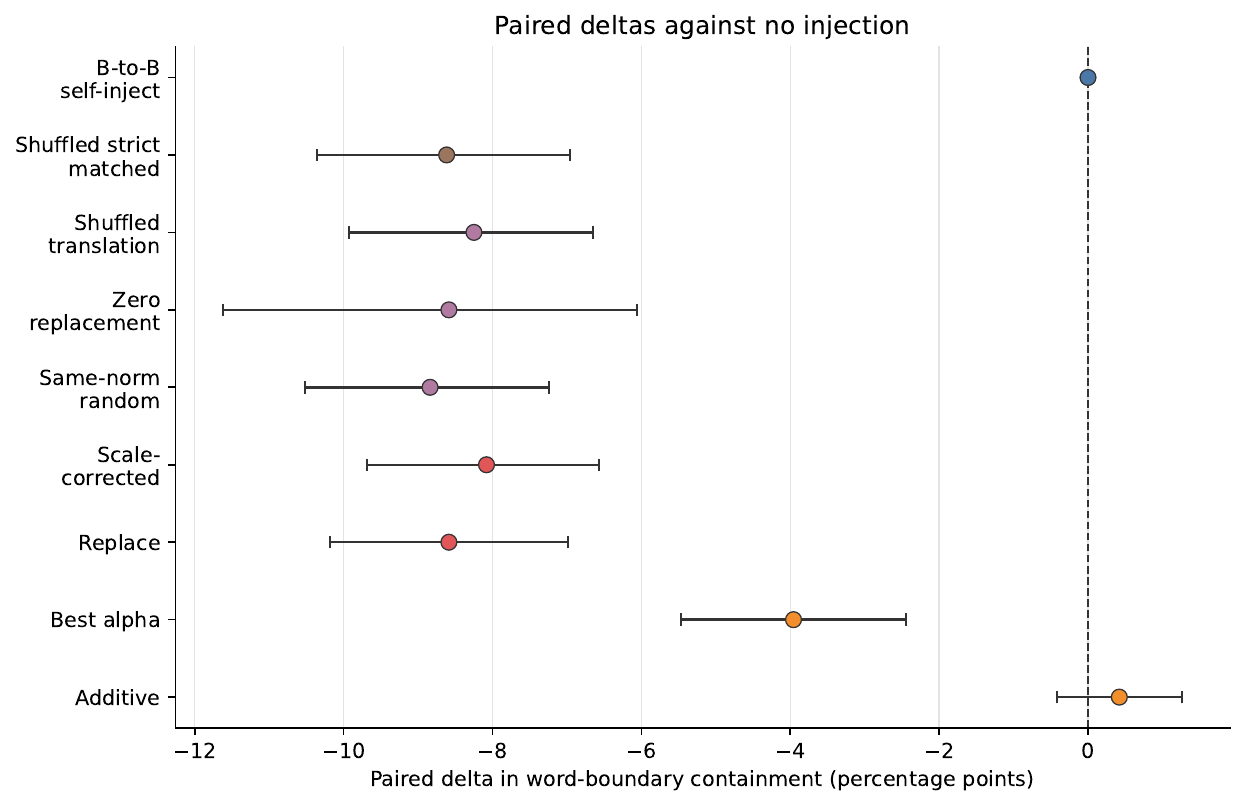}
\caption{Paired word-boundary deltas against no-injection. Additive injection
has a small positive mean but its pooled bootstrap confidence interval crosses
zero. Replacement-style conditions are negative with intervals below zero.}
\label{fig:delta-no-inject}
\end{figure}

\begin{figure}[H]
\centering
\includegraphics[width=0.88\linewidth]{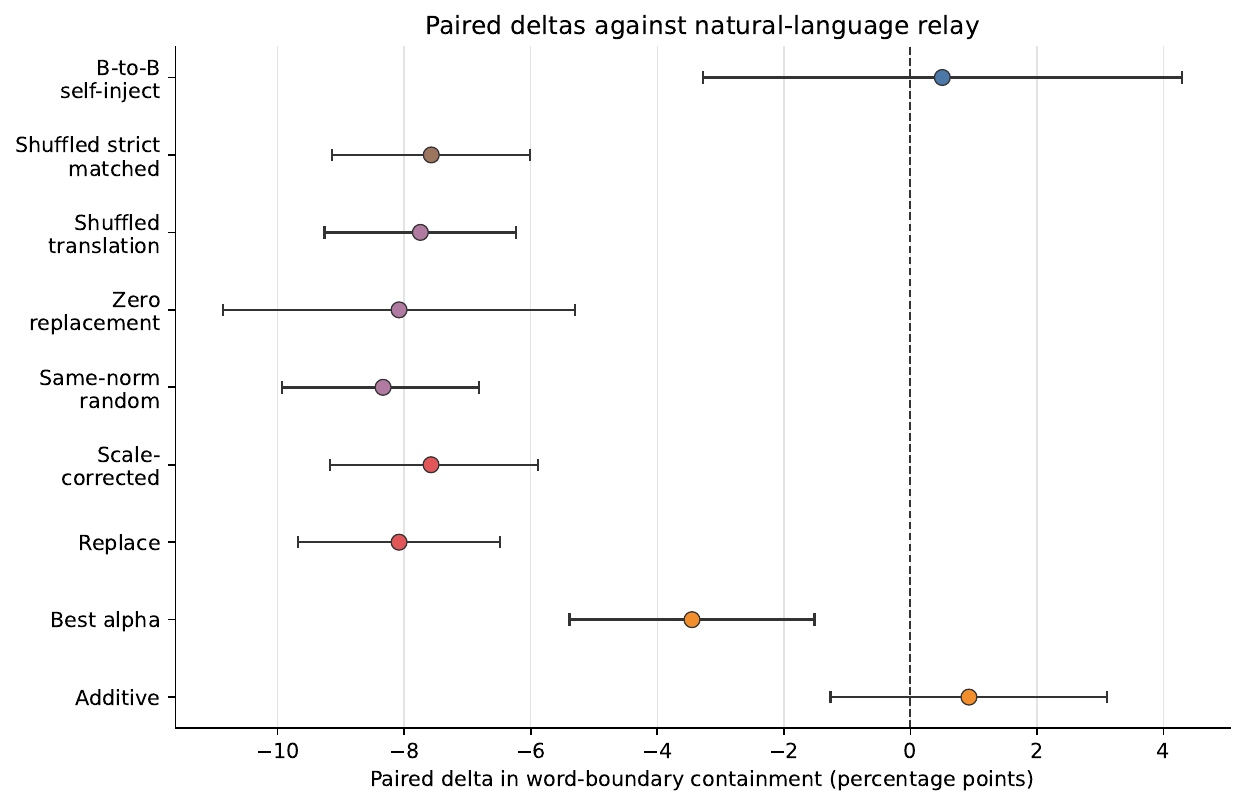}
\caption{Paired word-boundary deltas against natural-language relay. The
activation-transfer conditions do not provide a stable improvement over the
text communication baseline.}
\label{fig:delta-nl-relay}
\end{figure}

Relative to no-injection, additive has mean word-boundary delta $+0.0042$ with
pooled 95\% CI $[-0.0042,+0.0126]$. Relative to natural-language relay,
additive has mean delta $+0.0093$ with pooled 95\% CI $[-0.0126,+0.0311]$.
Both intervals cross zero. By contrast, replacement has delta $-0.0859$ versus
no-injection and $-0.0808$ versus natural-language relay, and scale-corrected
replacement remains strongly negative.
Appendix~\ref{app:pairwise} reports the compact paired-comparison table for
all word-boundary conditions.

\subsection{Failure Analysis}
\label{sec:failure}

The translation layer learns a stable normalized-space map, so the failure is
not a failure to fit hidden states. The downstream result instead shows that
hidden-state similarity under an offline loss does not imply that the receiver
can causally use the translated state inside its residual stream.

\begin{figure}[H]
\centering
\includegraphics[width=0.84\linewidth]{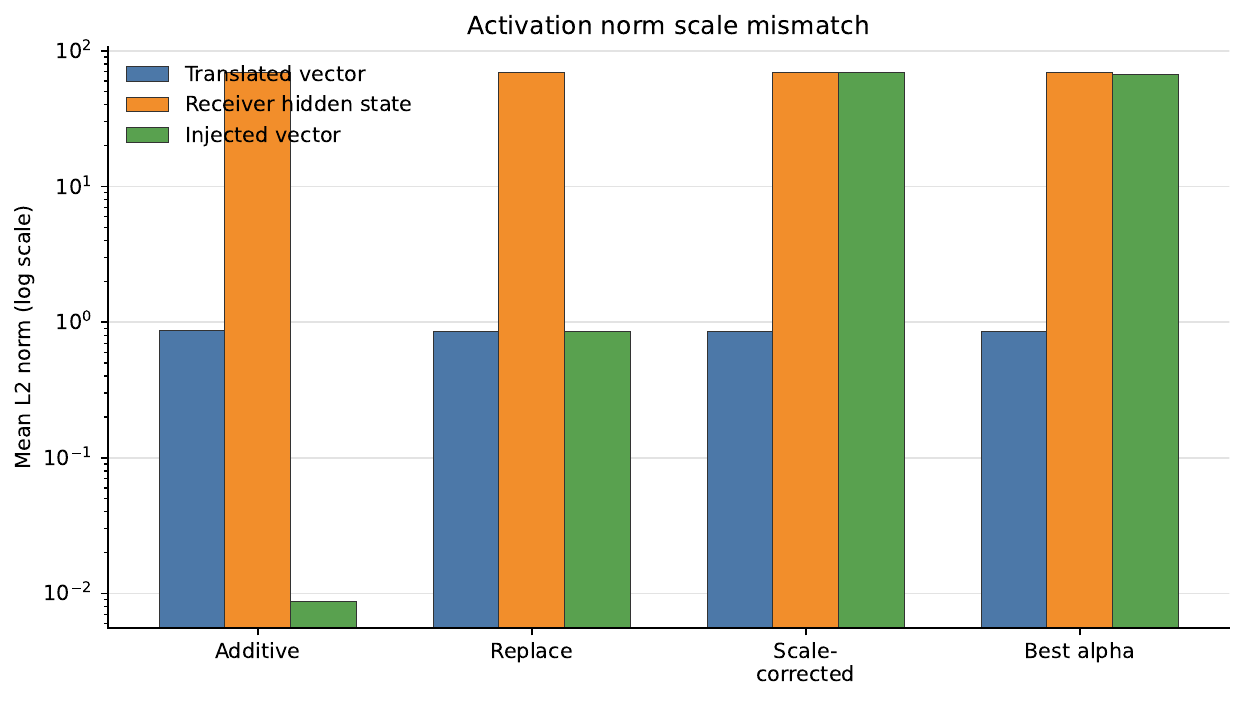}
\caption{Activation norm scale mismatch. Uncorrected translated vectors have
much smaller norm than native receiver hidden states. Scale correction removes
this simple mismatch but does not recover baseline behavior.}
\label{fig:norm-mismatch}
\end{figure}

Uncorrected replacement inserts translated vectors with mean norm around 0.85
where the receiver's native hidden-state norm is about 68.70. This severe norm
mismatch explains why uncorrected replacement is destructive. However,
scale-corrected replacement remains far below baseline, so norm mismatch is
not the whole explanation.

The injection conditions form an ablation chain that separates the two
failure factors. Additive injection leaves the residual stream nearly intact
and tracks the no-injection baseline; the best-alpha blend partially
substitutes the translated state and falls roughly halfway between baseline
and the strong-replacement floor; scale-corrected replacement removes the
L2-norm gap but otherwise substitutes the receiver state; and uncorrected
replacement combines norm mismatch and direction or distribution error. The gap between
uncorrected and scale-corrected replacement quantifies the contribution of
L2-norm mismatch, which is small relative to the gap between scale-corrected
replacement and no-injection. The latter gap---over eight percentage points
of word-boundary accuracy---persists after the norm correction. Under this
ablation, the remaining gap is consistent with residual direction or
distribution error being an important limiting factor, while the L2-norm gap
contributes only marginally.

\subsection{Diagnostic Controls}
\label{sec:controls}

The control set follows a single rule: each control isolates exactly one
rival explanation. The B-to-B self-injection condition routes the receiver's
own native activations through the same hook path; it exactly matches
no-injection on word-boundary containment ($0.0884$), which establishes that
the hook mechanism preserves behavior when the inserted states are native
receiver states. The same-norm random control injects receiver-scale random
vectors and drops to zero word-boundary accuracy, ruling out the explanation
that any vector of correct magnitude suffices. The full-n shuffled translation
condition pairs translated states with the wrong samples and is destructive
at the same level as replacement, showing that the sample identity of the
translated state carries the residual signal rather than any generic
translated-vector property. The strict matched shuffled control evaluates on
the 383-sample subset that excludes singleton prompt-length buckets, removing
the self-fallback caveat from the full-n shuffled diagnostic, and remains
destructive at the same level.

The final clean-eval validation report passes all blocking checks: 21 of 21
expected main result files are present, each main result file has $n=396$,
the clean-eval hash is consistent, and sequence and token mismatch counts are
zero. The combined validation report with strict matched shuffled controls
also passes blocking checks with 24 of 24 expected files; the strict matched
shuffled runs use the expected $n=383$ subset and have zero shuffle
self-fallback. Appendix~\ref{app:validation} reports the full validation
state, including the 13 self-fallback samples retained as a disclosed caveat
in the full-n shuffled translation files.

\section{Discussion and Limitations}

The evidence supports a narrow conclusion: in the tested setting, a translated
activation vector that is close under an offline normalized-space objective is
not automatically a useful communication object for the receiver. The controls
sharpen this further. B-to-B self-injection shows that the hook mechanism can
preserve behavior. Same-norm random and shuffled translation show that merely
injecting receiver-scale vectors is not sufficient. Offline alignment under an
L2-normalized loss and receiver-side causal usability are therefore separable
properties: the first is achieved here with normalized cosine similarity near
$0.97$ and normalized $R^2$ near $0.88$, while the second is not achieved at
any tested injection strength.

The negative result is scoped. It covers one model family, one model pair,
one primary direction, one task family, and one fixed injection depth. The
receiver is not adapted to consume translated activations. The models are raw
causal language models, and the metrics are automatic. Baseline word-boundary
accuracy is below $10\%$, so the negative result speaks to a low-baseline,
weak-receiver regime; stronger receivers, easier tasks, or receiver-adapted
transfer protocols could in principle change the picture. The best-alpha
condition uses $\alpha=0.30$ from a historical pilot that predates the
clean-eval final rerun. The strict matched shuffled control uses $n=383$
rather than the full $n=396$ because singleton prompt-length buckets are
excluded.

The negative result should not be read as evidence against hidden information
transfer between models. Subliminal learning shows that such transfer can occur
through a data-mediated, training-time channel \citep{cloud2026subliminal}; our
finding is that this does not automatically extend to inference-time linear
activation injection. The result therefore identifies a boundary condition for
this activation-mediated variant: offline linear alignment of a hidden state
does not by itself make that state causally usable when injected into a frozen
receiver. The base-model matching requirement reported for subliminal learning
is consistent with this interpretation, although the mechanism and channel are
different. A success obtained by distilling one model's outputs into another
therefore cannot be taken to imply that translating and injecting hidden
activations will succeed as well; offline representational alignment is not the
same as receiver-side causal usability.

These limitations directly motivate the constructive interpretation. The
experiment narrows the design problem for future work: useful model-to-model
activation communication likely requires a training objective that optimizes
receiver-side causal use, not only sender-to-receiver representational fit. Our
results caution against assuming that independently trained LLM agents can
replace text or structured-message relays with this kind of post-hoc linear
activation bridge. In the tested setting, offline alignment alone was not
receiver-usable. For related model-editing and activation-engineering
pipelines, the result suggests a practical caution: replacing receiver-side
activations may need to address both scale mismatch and residual direction or
distribution error. In this experiment, the translated vectors initially have a
large L2-norm gap---roughly two orders of magnitude relative to the receiver's
native hidden state---and accuracy remains far below baseline even after scale
correction closes that gap. This suggests that residual direction or
distribution error remains an important limiting factor for this bridge under
this protocol. This is a narrow statement about this bridge under this protocol
and is not a claim about covert or latent-channel communication between models
in general.

\section{Conclusion}

The clean rerun finds no evidence that the tested cross-model
activation-transfer mechanism improves multi-hop answering. Translation
quality is high in normalized space, but downstream injection fails.
Replacement-style transfer is harmful, and scale correction does not rescue
it. The main lesson is that causal usability inside the receiver is a
separate requirement from offline representational alignment, and future work
on model-to-model activation communication should optimize for receiver-side
causal use rather than for sender-to-receiver representational fit alone.

\section*{Author Contributions}

P.Z. and J.X. contributed equally to this work. J.X. proposed the research
idea and designed the experiments. P.Z. implemented and ran the experiments,
performed the data analysis, and finalized the manuscript.

\section*{Code and Data Availability}

The code, clean evaluation data, validated result files, and reproduction
commands for this study are available at
\url{https://github.com/jaysinailabs/cross-model-activation-transfer}.

\appendix

\section{Protocol Details}
\label{app:protocol}

The final rerun was executed under a frozen protocol rather than by selecting
favorable historical runs. Table~\ref{tab:frozen-defaults} summarizes the
defaults most likely to affect interpretation. The historical best-alpha value
is included because it is a possible source of confusion: it was selected from
an older pilot before the clean-eval final rerun and was then frozen.

\begin{table}[H]
\centering
\small
\begin{tabular}{ll}
\toprule
Item & Frozen value \\
\midrule
Primary direction & Pythia-160M to Pythia-410M \\
Task & Multi-hop reasoning \\
Clean-eval size & 396 samples \\
Sender layer & \texttt{gpt\_neox.layers.8} \\
Receiver layer & \texttt{gpt\_neox.layers.16} \\
Receiver decoding & Greedy, \texttt{max\_new\_tokens=64} \\
Natural-language relay length & \texttt{max\_new\_tokens=128} \\
Additive injection scale & 0.01 \\
Best-alpha value & 0.30, frozen from historical pilot \\
Main seeds & 42, 123, 456 for stochastic-condition replication \\
Strict shuffled subset & 383 samples, no self-fallback \\
\bottomrule
\end{tabular}
\caption{Protocol defaults fixed before interpreting the final clean rerun.}
\label{tab:frozen-defaults}
\end{table}

\section{Validation State}
\label{app:validation}

The results in this paper are produced by a frozen protocol together with a
clean-eval builder, a final runner, summary scripts, a paired comparison
script, a validation script, and a figure generation script, which emit the
result JSON files, validation reports, and audit material referenced below.
The final validation scripts check result completeness, sample count, clean
hash consistency, required JSON fields, sequence mismatch diagnostics, token
mismatch diagnostics, and strict shuffled-control invariants. The current
state is:

\begin{itemize}
\item Main final directory: 21 of 21 expected files present, blocking checks
pass.
\item Final directory plus strict matched controls: 24 of 24 expected files
present, blocking checks pass.
\item Clean-eval rebuild test: row count 396, acceptance flag true, and frozen
SHA-256 hash reproduced.
\item Full test suite: 79 tests passing at the time this draft was created.
\end{itemize}

The full-n shuffled translation files still report 13 self-fallback samples
per seed because 13 prompt-length buckets contain only one sample. This is
retained as a disclosed diagnostic caveat. The strict matched shuffled files
are the no-self shuffled control used for the paper-primary interpretation.

\section{Full Pairwise Word-Boundary Comparisons}
\label{app:pairwise}

Table~\ref{tab:pairwise-word-boundary} gives the paired word-boundary
comparison values used in the main result interpretation. Values are
percentage-point deltas with pooled bootstrap 95\% confidence intervals. The
same paired sample identifiers are used within each comparison. The strict
matched shuffled control has a smaller pooled count because it uses the
383-sample matched subset.

\begin{table}[H]
\centering
\small
\resizebox{\linewidth}{!}{%
\begin{tabular}{lrrr}
\toprule
Condition & vs. no injection & vs. natural-language relay & Pooled pairs \\
\midrule
Additive & +0.42 [-0.42, +1.26] & +0.93 [-1.26, +3.11] & 1188 \\
Best alpha & -3.96 [-5.47, -2.44] & -3.45 [-5.39, -1.52] & 1188 \\
Replace & -8.59 [-10.19, -6.99] & -8.08 [-9.68, -6.48] & 1188 \\
Scale corrected & -8.08 [-9.68, -6.57] & -7.58 [-9.18, -5.89] & 1188 \\
Same-norm random & -8.84 [-10.52, -7.24] & -8.33 [-9.93, -6.82] & 1188 \\
Zero replacement & -8.59 [-11.62, -6.06] & -8.08 [-10.86, -5.30] & 396 \\
Shuffled translation & -8.25 [-9.93, -6.65] & -7.74 [-9.26, -6.23] & 1188 \\
Strict matched shuffled & -8.62 [-10.36, -6.96] & -7.57 [-9.14, -6.01] & 1149 \\
B-to-B self-inject & +0.00 [+0.00, +0.00] & +0.51 [-3.28, +4.29] & 396 \\
\bottomrule
\end{tabular}
}
\caption{Paired word-boundary comparison summary in percentage points.}
\label{tab:pairwise-word-boundary}
\end{table}

\section{Metric Disagreement Audit}
\label{app:metric-audit}

\begin{figure}[H]
\centering
\includegraphics[width=0.72\linewidth]{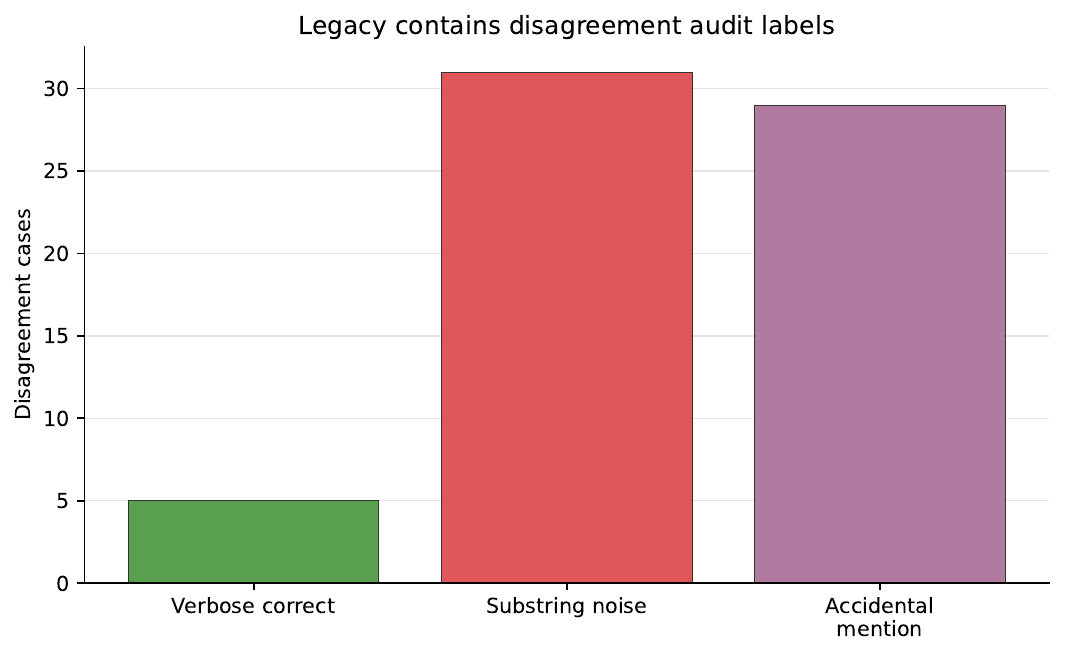}
\caption{Rule-derived labels for cases where legacy contains is true but
normalized exact match is false. The audit supports using word-boundary
contains as the main lenient metric.}
\label{fig:metric-audit}
\end{figure}

The metric-disagreement audit focuses on rows where legacy contains-match is
true while normalized exact match is false. There are 65 such rows in the
audit slice. Figure~\ref{fig:metric-audit} shows the rule-derived label
distribution: 5 verbose-correct cases (all from natural-language relay), 31
substring-noise cases where the legacy substring rule fires for a non-answer
substring, and 29 accidental-mention cases where the gold answer appears in
the prediction but not as the direct answer. A 10-row non-scripted spot-check
covering all three labels agrees with the rule-derived labels.

This audit is why the main text uses word-boundary containment as the lenient
metric and treats legacy contains-match as historical context.

\section{Reproducibility Commands}
\label{app:commands}

The following commands summarize the workflow that generated the results in
this paper, run from the project root.

\begin{verbatim}
python papers/h1_activation_transfer/scripts/build_clean_eval.py
python papers/h1_activation_transfer/scripts/check_readiness.py \
  --require-model-cache
python papers/h1_activation_transfer/scripts/summarize_results.py
python papers/h1_activation_transfer/scripts/compare_results.py
python papers/h1_activation_transfer/scripts/build_metric_disagreement_audit.py \
  --per-result-file 5
python papers/h1_activation_transfer/scripts/validate_final_results.py
python papers/h1_activation_transfer/scripts/generate_figures.py
pytest tests
\end{verbatim}

The strict matched shuffled controls are validated together with the main final
directory using:

\begin{verbatim}
python papers/h1_activation_transfer/scripts/validate_final_results.py \
  --results-dir papers/h1_activation_transfer/results/final \
                papers/h1_activation_transfer/results/final_strict_controls \
  --include-strict-matched \
  --output-dir \
    papers/h1_activation_transfer/results/final_with_strict_controls
\end{verbatim}

\bibliographystyle{plainnat}
\bibliography{references}

\end{document}